\pdfoutput=1
\documentclass[twoside]{article} 
\usepackage{aistats2017}

\usepackage[utf8]{inputenc} 
\usepackage{hyperref}       
\usepackage{url}            
\usepackage{booktabs}       
\usepackage{nicefrac}       
\usepackage{microtype}      
\usepackage{amsmath}
\usepackage{amssymb}
\usepackage{graphics} 
\usepackage{times}    
\usepackage{color}
\usepackage{textcomp}
\usepackage{comment,subcaption,multirow,array,graphicx}
\usepackage{algorithmicx}
\usepackage{algorithm}
\usepackage[noend]{algpseudocode}
\usepackage{mathtools}

\usepackage[nottoc,notlof,notlot]{tocbibind} 
\usepackage[titles,subfigure]{tocloft} 

\usepackage{bm}

\newcommand{\va}{\bm{a}}               
\newcommand{\vb}{\bm{b}}

\newcommand{\vo}{\bm{o}}

\newcommand{\vt}{\bm{t}}

\newcommand{\vx}{\bm{x}}               
\newcommand{\vy}{\bm{y}}               
\newcommand{\vz}{\bm{z}}

    \newcommand{\Dc}{\mathcal{D}}

    \newcommand{\Lc}{\mathcal{L}}  
    
    \newcommand{\Nc}{\mathcal{N}}  
    \newcommand{\Oc}{\mathcal{O}}

    \newcommand{\Yc}{\mathcal{Y}}

\begin{document}

\twocolumn[

  \aistatstitle{Learning Time Series Detection Models from Temporally
  Imprecise Labels}

  \aistatsauthor{Roy J. Adams$^\star$ \and Benjamin M. Marlin}

  \aistatsaddress{University of Massachusetts, Amherst} 

]

\begin{abstract}
In this paper, we consider the problem of learning time series detection models from temporally imprecise labels. In this problem, the data consist of a set of input time series, and supervision is provided by a sequence of noisy time stamps corresponding to the occurrence of positive class events. Such temporally imprecise labels occur in areas like mobile health research when human annotators are tasked with labeling the occurrence of very short duration events. We propose a general learning framework for this problem that can accommodate different base classifiers and noise models. We present results on real mobile health data showing that the proposed framework significantly outperforms a number of alternatives including assuming that the label time stamps are noise-free, transforming the problem into the multiple instance learning framework, and learning on labels that were manually aligned.
\end{abstract}

\section{Introduction}

A key problem with supervised learning is that the cost of obtaining accurate labeled data is often prohibitive. This problem has been addressed through a number of alternative learning frameworks where labels can be obtained at lower cost, including frameworks for learning from lower volumes of labeled data (semi-supervised learning \cite{chapelle2006semi}, positive unlabeled learning \cite{lee2003learning}, active learning \cite{settles2010active}, etc.), and frameworks for learning from lower-quality labels (multiple instance learning \cite{maron1998framework}, learning with label proportions \cite{hernandez-pr2013} etc.).

In this paper, we consider a new low-quality label learning problem: learning time series detection models from temporally imprecise labels. In this problem, we begin with a  multivariate time series and assume that it has been temporally discretized and featurized to produce a temporal sequence of instances. The discretization can be performed using sliding windows, unsupervised segmentation, or any other approach. Supervision is provided in the form of a sequence of noisy time stamps corresponding to the occurrence of positive class events. The primary assumption, which will be formalized in the proposed model, is that each observed positive event time stamp is generated probabilistically from a distribution centered close in time to a true positive instance. We subsequently relax this assumption to allow for both false positive event time stamps and missed event time stamps. The goal is to use this data to learn an instance-level classifier that accepts an instance as input and outputs the correct class label.

This problem is motivated by research in the area of mobile health (or mHealth) \cite{kumar2013mobile} where human annotators are often tasked with labeling the occurrence of very short duration events during lab or field-based data collection studies. For example, Ali et al. considered the problem of learning classification models to detect the occurrence of individual smoking puffs based on data from a respiration chest band sensor \cite{ali2012mpuff}. While a subject instrumented with a chest band sensor smoked a cigarette, the annotator observed the subject and pressed a button on a smartphone to record a time stamp for each smoking puff. The sensor data time series was then segmented into respiration cycles and features were extracted from each cycle. 

As reported by Ali et al. the observer-supplied puff time stamps were sufficiently imprecise that they had to be manually aligned with the respiration cycles by visualizing the 20Hz respiration waveforms before models could be usefully learned \cite{ali2012mpuff}. This process required significant time and has the potential to result in annotation errors. This same data collection procedure and post hoc label alignment has been used in more recent work on smoking detection \cite{saleheen-ubicomp2015}, and similar problems occur in other mHealth research areas including learning to detect eating and drinking gestures from wrist-worn actigraphy sensor data streams \cite{thomaz-ubicomp2015}.

To address this problem, we propose a modeling framework for learning detection models directly from temporally imprecise labels without discarding any available information. The framework can be used to learn the parameters of a discriminative probabilistic classifier that requires independent instances and fully observed labels during learning. Our framework augments a chosen base classifier with a probabilistic model of the positive event time stamp generation process to account for temporally noisy labels, as well as a label value noise process to account for the possibility of false positives and negatives in the event labels. 

We provide a gradient-based algorithm for jointly  learning all model parameters by maximizing the marginal likelihood of the observed positive label time stamps given the sequence of instances features. The learning algorithm leverages a dynamic programming-based exact inference algorithm that requires $O(L\cdot M^2)$ time where $L$ is the number of instances and $M$ is the number of positive event time stamps. The learned base classifier is then used alone to to classify individual instances at test time, as if it had been trained on standard fully observed labels. 

We present results on real mobile health data showing that the proposed framework significantly outperforms a number of alternatives including assuming that the label time stamps are noise-free, transforming the problem into the multiple instance learning framework, and learning on labels that were manually re-aligned.

\section{Related Work}

In this section, we briefly review related work on alternative frameworks for reducing the cost of collecting labels for supervised learning. The most obvious way to decrease the cost of obtaining labeled data is to decrease the amount of data that needs to be labeled. \textit{Active learning} aims to optimally select the instances that are labeled from an unlabeled pool or stream of instances with the goal of learning better models from lower volumes of labeled data \cite{settles2010active}. \textit{Semi-supervised learning} aims to learn from small volumes of labeled data by combining it with large amounts of unlabeled data \cite{chapelle2006semi}. \textit{Positive unlabeled learning} is a special case of semi-supervised learning where a small number of positive instances have labels, but a large pool of unlabeled instances are available \cite{lee2003learning}. 


An alternative to labeling less data is to lower the cost of obtaining each label, which can be achieved by lowering the quality of labels in some way. For example, \textit{multiple instance learning} generalizes supervised learning by allowing for sets (or ``bags'') of instances to be labeled instead of single instances \cite{maron1998framework}. The label semantics require that a bag be labeled positive if it contains at least one positive instance and that it be labeled negative otherwise.  The closely related \textit{label proportions} framework provides the proportion of each type of label for a group of instances \cite{hernandez-pr2013}.

Hern{\'a}ndez-Gonz{\'a}lez at al. provide a detailed categorization of existing non-standard classification problems in terms of their instance-label relationships and supervision frameworks \cite{hernandez-prl2016}. They describe four possible instance-label relationships including \textit{single-instance single-label} (SISL), \textit{single-instance multiple-label} (SIML), \textit{multiple-instance single-label} (MISL), and \textit{multiple-instance multiple-label} (MIML). They describe a number of supervision frameworks including full supervision, semi-supervision, positive-unlabeled, noisy label values, probabilistic labels, and label proportions, among others.

We note that the problem we consider here, temporally noisy labels, does not fit cleanly into the existing categorization of methods provided by Hern{\'a}ndez-Gonz{\'a}lez et al. The form of the classifier that we wish to induce is single-instance single-label; however, we do not know the correct labels for any instances. Instead, we use the intuition that positive event labels are recorded ``close'' in time to true positive instances (this intuition will be formalized in the following section). Our problem can be converted to the full supervised setting by assuming that the positive event labels should be assigned to the temporally closest instance and that all other instance labels should be negative. This transformation will convert noise in the time stamps of the positive event labels into noise in the values of the instance labels. We will refer to this as the \textit{naive supervision} strategy. 

A potentially more robust transformation of our problem is to partition the instances into temporally contiguous, non-overlapping bags and to label the bags positive if there exists a positive event label within the time span that defines the bag. Multiple instance learning methods can then be applied. This transformation may improve on the naive supervision strategy, but temporally noisy labels can still spill across adjacent bags, resulting in noise in the values of the bag labels. Further, this transformation discards information about the number of positive event labels that occur within a bag. Finally, the label proportions framework can be applied by again switching to a temporal bag-based approach, but labeling the bags with the number of positive event labels within the time span that defines the bag. This approach still discards information about the observed locations of the positive event labels, which the approach we propose fully retains. Furthermore, we extend the basic temporally imprecise label problem to allow for both false positive event labels and missed positive event labels. 

In terms of model structure, our framework resembles approaches used for learning classifiers in the presence of label noise \cite{yan2010modeling,yu2010modeling,raykar2009supervised,jin2002learning}. Each of these papers proposes a version of a latent variable model where the true instance label is assumed to be unobserved, but a noisy version of the label value is observed. The learning algorithms and observation noise models depend on the specific applications, but all involve maximizing the marginal likelihood of the observed noisy label values. These problems differ from ours in that they assume a noisy label value is provided for each instance independently, whereas we assume that a sequence of temporally noisy labels is given for a sequence of instances.

Another closely related model is the Connectionist Temporal Classification (CTC) model proposed by Graves et al. \cite{graves2006connectionist}. This model estimates the weights of a recurrent neural network (RNN) from unaligned feature and label sequences. Similar to our approach, alignment of the two sequences is treated as a latent variable that is marginalized out; however, Graves et al. do not explicitly model the observation noise distribution\cite{graves2006connectionist}. In fact, the CTC model is a special case of our proposed framework that uses a RNN as the base classifier, a binary observation count distribution, and a uniform timestamp noise distribution.

\vspace{-1em}
\section{Model, Inference, and Learning}

In this section, we present the proposed modeling framework for learning time series detection models from temporally imprecise labels. We begin by introducing the required notation to define the problem, and then proceed to defining the model and deriving inference and learning algorithms.

\textbf{Problem Description and Notation:} We assume that the available data are organized into $N$ sessions. Each session $n$ contains a multivariate time series and an event label sequence. The different sessions may be generated by observing the same processes at different times, or, in the mHealth case, by collecting data from different subjects. We assume that the time series for each session has been discretized into a sequence of $L_n$ potentially irregular and/or overlapping sub-windows, and that a feature vector $\vx_{ni} \in \mathbb{R}^{D}$ has been extracted from each window $i$ in each session $n$. We refer to $\vx_n=[\vx_{n1},...,\vx_{nL_n}]$ as the instance sequence associated with session $n$. Each instance is associated with a time point $t_{ni}$ that may denote the start, end, or other point of interest in the window the instance was derived from, depending on the discretization process. We define $\vt_n=[t_{n1},...,t_{nL_n}]$ to be the sequence of instance time points. 

Each instance also has a corresponding label variable $Y_{ni}\in \{0,1\}$. However, the instance labels $y_{ni}$ are not directly observed. Instead, we observe a sequence of $M_n$ event time stamp values $z_{nm} \in \mathbb{R}_+$ indicating the times where positive events were recorded, and these time stamps are assumed to be noisy. We denote the sequence of event time stamps for session $n$ by $\vz_{n}=[z_{n1},...,z_{nM_n}]$, which we assume to be in increasing order without loss of generality. 

The problem we aim to solve is to learn a standard instance-level classification function $f: \mathbb{R}^{D} \rightarrow \{0,1\}$ from a data set $\mathcal{D}=\{(\vx_n,\vt_n,\vz_n)|1\leq n\leq N\}$ consisting of the instances, the instance time stamps, and the positive event time stamps. 

\textbf{Proposed Modeling Framework:} To solve the problem outlined above, we formalize the assumptions that the observed positive event time stamps should be ``close'' in time to true positive instances, and that false negative and false positive time stamps are possible using an explicit generative model. Our proposed framework includes three components that can be chosen independently: a probabilistic base classifier, an observation count distribution, and an observation timestamp noise distribution. 

For the base classifier, we assume a differentiable discriminative classifier of the form $p_{\theta}(y_{ni}|\vx_{ni})$ with parameters $\theta$. Any discriminative probabilistic classifier where the label variables $y_{ni}$ are probabilistically independent given the features $\vx_{ni}$ can be used. Such models include logistic regression \cite{hosmer2004applied} and kernel logistic regression \cite{zhu2001kernel}, as well as multi-layer feedforward neural networks \cite{hornik1989multilayer}, convolutional neural networks \cite{krizhevsky2012imagenet}, and recurrent neural networks \cite{graves2013speech} when used with logistic output layers.

Our goal can now be stated as the problem of accurately estimating the parameters $\theta$ of the base classifier $p_{\theta}(y_{ni}|\vx_{ni})$ from $\Dc= \{(\vx_n, \vt_n, \vz_n)|1\leq n\leq N\}$. To accomplish this learning task, we augment the base classifier with a generative model of the observation process described below where we have introduced a latent count variable $o_{ni} \in \{0,...,M_n\}$ for each instance $i$ that represents the number of positive event time stamps generated by that instance.
\begin{enumerate}
	\item Given an instance label $y_{ni}$, the number of observed positive event time stamps generated by this instance is drawn according to $o_{ni} \sim p_\pi(o_{ni}|y_{ni})$.
	\item $o_{ni}$ observation timestamps are then drawn independently according to $z_{nm} \sim p_\phi(z_{nm}|t_{ni})$.
\end{enumerate}
	
We define $p_\pi(o_{ni}|y_{ni})$ as the distribution over the number of observations generated by an instance with label $y_{ni}$. We can choose any differentiable, parametric form for $p_\pi(o_{ni}|y_{ni})$. A simple example we will use later is $p_\pi(o_{ni} = 1|y_{ni}) = 1 - p_\pi(o_{ni} = 0|y_{ni}) = \pi_{y_{ni}}$ for $\pi_v \in [0,1]$, which encodes a model where true positives may be missed, and there can be at most one false observation per negative instance.

The conditional probability density $p_{\phi}(z_{nm}|t_{ni})$ generates the event time stamp $z_{nm}$ given the corresponding instance time stamp $t_{ni}$. As above, we can choose any parametric form for the density function $p_{\phi}(z_{nm}|t_{ni})$. In this work, we focus specifically on the case where $p_{\phi}(z_{nm}|t_{ni})$ is the conditional normal distribution $\Nc(z| \beta + t_{ni}, \sigma^2)$ where $\beta$ provides a time stamp bias parameter relative to the instance time stamp, and $\sigma^2$ is the variance of the noise distribution. Many other choices are possible within the framework including different ways of conditioning the mean and variance of $p_{\phi}(z_{nm}|t_{ni})$ on $t_{ni}$, as well as the use of completely different distributions for $p_{\phi}(z_{nm}|t_{ni})$, such as skewed distributions like the Gamma distribution.

A consequence of this generative process is that the posterior support set for $\vo_{n}=[o_{n1},..,o_{nL_n}]$ consists of all vectors with a sum equal to $M_n$. We denote this set by $\Oc_{n} = \left\{\vo| \vo \in \{0,1\}^{L_n} , \sum_{i=1}^{L_n} o_{ni}=M_n\right\}$. For presentation, we will also denote the set of all possible label vectors $\vy_{n}$ by $\Yc_{n} = \{\vy | \vy \in \{0,1\}^{L_n}\}$.

At this point, we have defined the base noise distribution $p_{\phi}(z_{nm}|t_{ni})$, which defines the conditional probability density of an event time stamp given the time stamps for the corresponding positively labeled instance; however, in the observed data, we do not know the correspondence between instances and event time stamps. To solve this problem, we introduce the following mild simplifying assumption: \textit{If instance $j$ occurs after instance $i$ in the input sequence, then any positive event time stamps generated by instance $j$ must occur after any positive event time stamp generated by instance $i$.}

We define the indicator function $w(i,m) = \sum_{j=1}^{i-1} o_{nj} < m \leq \sum_{j=1}^{i} o_{nj}$, which returns $1$ when $z_{nm}$ was generated by instance $i$, and $0$ otherwise. We can now express the above simplifying assumption mathematically as the condition: if $i < j$, $w(i,k)=1$ and $w(j,l)=1$ then $z_{nk} < z_{nl}$. 



We specify the joint distribution over the event time stamp variables $\vz_n$ given the observation count variables $\vo_n$ and instance time stamp variables $\vt_n$ in Equation \ref{eq_pzgo}.\footnote{The notation $[s]$ is the Iverson bracket where $[s]=1$ if the statement $s$ is true and $[s]=0$ if the statement $s$ is false.} 
\begin{align}
\label{eq_pzgo}   
        &p(\vz_n|\vo_n,\vt_n) = \prod_{i=1}^{L_n}
        \left(\prod_{l=1}^{M_n} 
          p_{\phi}(z_{nl}|t_{ni})^{w(i,l)}
        \right)^{[o_{ni}>0]}
\end{align}
The complete joint model for the instance labels and event time stamps given the feature variables, and instance time stamps is shown in Equation \ref{eq_pyz} where $\psi = \{\theta,\phi,\pi\}$ is the complete set of model parameters. 
\begin{align}
\nonumber
        p_{\psi}(\vy_n,&\vo_n,\vz_n|\vx_n,\vt_n) \\
        &= p_{\theta}(\vy_n|\vx_n)\cdot p_\pi(\vo_{n}|\vy_{n})\cdot p_{\phi}(\vz_n|\vo_n,\vt_n)
\label{eq_pyz}   
\end{align}
The instance label and observation count variables $\vy_n$ and $\vo_n$ are not observed during learning, but we can marginalize them out of the model as seen in Equation \ref{eq_pzgx}. The marginalization operation is expressed in terms of a sum over the support sets $\Oc_n$ and $\Yc_n$ of $\vo_n$ and $\vy_n$ respectively. In the next sections, we derive efficient inference and learning algorithms for maximizing the marginalized conditional likelihood of the data. We note again that once the model is learned, the observation count distribution and the timestamp noise distribution can be discarded and instances can be classified using $p_{\theta}(y_{n}|\vx_{n}).$
\begin{align}       
\label{eq_pzgx} 
        p_{\psi}(\vz_n|\vx_n, \vt_n) &=
        \sum_{\vy_n \in \Yc_{n} }\sum_{\vo_n \in \Oc_{n} }
        p_{\psi}(\vy_n,\vo_n,\vz_n|\vx_n,\vt_n)
\end{align}
\begin{figure*}
\begin{align}
\label{eq:a_def}
&a(i,l)  =  p_{\psi}\left(\vz_{1:l}\left|{\textstyle \sum_{j=1}^i o_j=l},\vx_{1:i},\vt_{1:i}\right.\right)
 = 
\sum_{\substack{\vy_{1:i}\in \Yc_{i}\\ \vo_{1:i}\in \Oc_{i,l}}}
p_{\psi}(\vy_{1:i},\vo_{1:i},\vz_{1:l}|\vx_{1:i},\vt_{1:i})\\
%
\label{eq:b_def}
&b(i,l)  =  p_{\psi}\left(\vz_{l:M}\left|{\textstyle \sum_{j=i}^{M}o_j=l},\vx_{i:L},\vt_{i:L}\right.\right) \hspace{-0.25em} = \hspace{-1em}
\sum_{\substack{\vy_{i:L}\in \Yc_{L-i}\\ \vo_{i:L}\in \Oc_{L-i,M-l}}}
p_{\psi}(\vy_{i:L},\vo_{i:L},\vz_{l:M}|\vx_{i:L},\vt_{i:L}) \\[20pt]
\label{eq:py_zxt}
&p_{\psi}(y_i|\vz,\vx,\vt)
    =  \sum_{c=0}^{M} \sum_{j=1}^{M-c} \frac{a(i-1,j-1)b(i+1,j+c+1)\prod_{k=j}^{j+c}p_{\phi}(z_k|t_i)\cdot p_\pi(o_i=c|y_i)p_\theta(y_i|\vx_i)}{a(L,M)}\\
%
\label{eq:pw_zxt}
&p_{\psi}(w(i,l)|\vz,\vx,\vt) 
    =  \sum_{c=1}^{M} \sum_{j=l-c}^{l} \frac{a(i-1,j-1)b(i+1,j+c+1)\prod_{k=j}^{j+c}p_{\phi}(z_k|t_i)\cdot \sum_{y_i} p_\pi(o_i=c|y_i)p_\theta(y_i|\vx_i)}{a(L,M)}\\
%
\label{eq:po_zxt} 
&p_{\psi}(o_i=c,y_i|\vz,\vx,\vt) 
    = \sum_{j=1}^{M-c} \frac{a(i-1,j-1)b(i+1,j+c+1)\prod_{k=j}^{j+c}p_{\phi}(z_k|t_i)\cdot p_\pi(o_i=c|y_i)p_\theta(y_i|\vx_i)}{a(L,M)}
\end{align}

\rule{6.75in}{0.5pt}

\end{figure*}
\textbf{Learning:} To learn the proposed model, we will maximize the marginal log likelihood $\Lc(\theta,\phi,\pi|\Dc)=\Lc(\psi|\Dc)$ where $\theta$ are the base classification model parameters, $\phi$ are the observation noise model parameters, $\pi$ are the observation count model parameters, and $\Dc$ consists of the observed data for each session. This objective function is defined below.
\begin{align}
\nonumber
\Lc(&\psi|\Dc)  =
\sum_{n=1}^N 
\log p_{\psi}(\vz_n|\vx_n,\vt_n)\\
&=\sum_{n=1}^N 
\log\left( \sum_{\vy_n \in \Yc_n }\sum_{\vo_n \in \Oc_n } 
p_{\psi}(\vy_n,\vo_n,\vz_n|\vx_n,\vt_n)
\right)
\end{align}
We develop a gradient-based learning approach for the proposed model. The complete derivations for all gradients presented in this section are available in the supplemental materials. We first consider the gradient of $\log p_{\psi}(\vz_n|\vx_n,\vt_n)$ with respect to the detector parameters $\theta$. We drop the dependence on the session index $n$ for brevity and will make use of the log probability of the base classifier, which we will denote by $\ell_{\theta}(y|\vx) = \log p_{\theta}(y|\vx)$.
\begin{align}
        \nabla_\theta \log p_{\psi}(\vz|\vx,\vt) &=\sum_{i=1}^L\mathbb{E}_{p_{\psi}(y_i|\vz,\vx,\vt)} \nabla_\theta\ell_{\theta}(y_i|\vx_i)
\end{align}
This gradient has the form of a sum of expected gradients for individual instances, where the expectation is taken with respect to the marginal posterior distribution $p_{\psi}(y_i|\vz,\vx,\vt)$. Assuming that the gradient of the base classifier can be computed efficiently, the complexity of computing the gradient of the proposed likelihood with respect to $\theta$ depends only on the complexity of computing the posterior marginal $p_{\psi}(y_i|\vz,\vx,\vt)$. Next, we give the gradient with respect to the noise model parameters $\phi$ and define $\log p_{\phi}(z_l|t_i) = \ell_{\phi}(z_l|t_i)$. 
\begin{align}
        \nabla_\phi \log p_{\psi}(\vz|\vx,\vt)  &= \sum_{l=1}^M
        \mathbb{E}_{p_{\psi}(w(i,l)|\vz,\vx,\vt)}        
        \nabla_\phi\ell_{\phi}(z_l|t_i)         
\end{align}
This gradient is again a sum of expectations for individual event time stamps. In this case, the sum is over each observed time stamp and the expectation is taken with respect to the posterior marginal distribution $p_{\psi}(w(i,l)|\vz,\vx,\vt)$, which gives the probability that instance $i$ is responsible for generating event time stamp $l$. Assuming that the gradients of $\ell_{\phi}(z_l|t_i)$ can be computed efficiently, the complexity of the gradient computation depends on the complexity of computing the posterior marginal $p_{\psi}(w(i,l)|\vz,\vx,\vt)$. 

Finally, we give the gradient with respect to the observation count  parameters, $\pi$, letting $\log p_\pi(o_{i}|y_{i}) = \ell_\pi(o_i|y_i)$.

\begin{align}
        \nabla_\pi \log p_{\psi}(\vz|\vx,\vt)  &= \sum_{i=1}^L\mathbb{E}_{p_{\psi}(o_{i},y_i |\vz,\vx,\vt)}\nabla_\phi\ell_{\pi}(o_i|y_i)         
\end{align}
The gradient is again a sum of expected gradients. In this case, the expectation is taken with respect to the posterior marginal $p_{\psi}(o_{i},y_i |\vz,\vx,\vt)$ which gives the joint posterior distribution over the label for instance $i$ and number of observations generated by instance $i$.

The gradient system required to optimize all parameters of the complete joint distribution is shown below in Equations \ref{eq:dtheta}, \ref{eq:dphi}, and \ref{eq:dpi}. We next turn to the problem of efficiently computing the required posterior marginal distributions $p_{\psi}(y_{ni}|\vz_n,\vx_n,\vt_n)$, $p_{\psi}(w(i,l)|\vz,\vx,\vt)$, and $p_{\psi}(o_{i},y_i |\vz,\vx,\vt)$.
%
%
%
%
\begin{align}
		\label{eq:dtheta}
        \nabla_\theta \Lc(\psi|\Dc) &=
        \sum_{n=1}^N\sum_{i=1}^{L_n}
        \mathbb{E}_{p_{\psi}(y_i|\vz,\vx,\vt)} \nabla_\theta\ell_{\theta}(y_i|\vx_i)
        \\
\label{eq:dphi}
        \nabla_\phi \Lc(\psi|\Dc) &=
        \sum_{n=1}^N \sum_{l=1}^{M_n}
        \mathbb{E}_{p_{\psi}(w(i,l)|\vz,\vx,\vt)}        
        \nabla_\phi\ell_{\phi}(z_l|t_i) \\
\label{eq:dpi}
        \nabla_\pi \Lc(\psi|\Dc) &=
        \sum_{n=1}^N \sum_{i=1}^{L_n}
        \mathbb{E}_{p_{\psi}(o_{i},y_i |\vz,\vx,\vt)}\nabla_\phi\ell_{\pi}(o_i|y_i)
\end{align}

\textbf{Inference:} In this section, we present a dynamic programming algorithm for exact posterior marginal inference in the proposed model. We show that the same dynamic program enables the computation of each of the desired posterior marginal distributions: $p_{\psi}(y_{i}|\vz,\vx,\vt)$, $p_{\psi}(w(i,l)|\vz,\vx,\vt)$, and $p_{\psi}(o_{i},y_i |\vz,\vx,\vt)$. The dynamic program involves a forward recursion with variables $a(i,l)$ and a backward recursion with variables $b(i,l)$. For compactness, we introduce the notation $\vy_{i:j}=[y_i,y_{i+1},...,y_{j}]$ and $\vo_{i:j}=[o_i,o_{i+1},...,o_{j}]$ to indicate the sub-vectors of $\vy$ and $\vo$ starting at position $i$ and ending at position $j$. We also introduce the notation $\Oc_{i,l} = \{\vo |\vo\in\{0,1\}^i, \sum_{j=1}^i o_j=l\}$ and $\Yc_{i} = \{\vy |\vy\in\{0,1\}^i\}$ to indicate the set of label and count vectors of length $i$ that generated exactly $l$ observations. We define $a(i,l)$ and $b(i,l)$ in Equations \ref{eq:a_def} and \ref{eq:b_def}.

Intuitively, $a(i,l)$ gives the marginal probability of the first $l$ event time stamps given that the first $l$ observations were generated by the first $i$ instances. Likewise, $b(i,l)$ gives the marginal probability of the last $M-l$ time stamps assuming that the last $M-l$ time stamps were generated by the last $L-i$ instances. By these definitions, the marginal likelihood is given by $p_{\psi}(\vz|\vx,\vt) = a(L,M) = b(0,0)$.
The recursions for computing the dynamic programming variables are given in Algorithm \ref{alg_inf}. The complete set of $a(i,l)$ and $b(i,l)$ values can clearly be computed in $O(L\cdot M^2)$ time using this dynamic program.

\begin{algorithm*}[t]
  \caption{Posterior Inference Dynamic Program for Temporally Imprecise Labels}
  \label{alg_inf}
  
\begin{algorithmic}[1]  
\State Inputs: $\theta, \phi, \pi, \vx\in \mathbb{R}^{L\times D}$, $\vt\in \mathbb{R}^{L\times 2}$, $\vz\in \mathbb{R}^{M}$
\State Let $\va \in \mathbb{R}^{L\times M}$, $\vb \in \mathbb{R}^{L\times M}$,
$a(0,0) \leftarrow 1$,
$b(0,0) \leftarrow 1$
\For{$i = 1,...,L$}
\For{$l = 0,...,M$}
\State
$a(i,l) \leftarrow 
\displaystyle
\sum_{y\in \{0,1\}} \sum_{c = 0}^l p_{\theta}(y_i=y|\vx_i)p_\pi(o_i=c|y,\vx_i)a(i-1,l-c)\prod_{j=l-c}^l p_\phi(z_j|t_i)$ 
\EndFor
\EndFor
\For{$i = L,...,1$}
\For{$l = M,...,0$}
\State $b(i,l) \leftarrow
\displaystyle
\sum_{y\in \{0,1\}} \sum_{c = 0}^l p_{\theta}(y_i=y|\vx_i)p_\pi(o_i=c|y,\vx_i)b(i+1,l+c)\prod_{j=l}^{l+c} p_\phi(z_j|t_i)$
\EndFor
\EndFor
\State Return $\va$, $\vb$
\end{algorithmic}
\end{algorithm*}




Finally, Equations \ref{eq:py_zxt}, \ref{eq:pw_zxt}, and \ref{eq:po_zxt} show how to calculate $p_{\psi}(y_{i}|\vz,\vx,\vt)$, $p_{\psi}(w(i,l)|\vz,\vx,\vt)$, and $p_{\psi}(o_{i},y_i |\vz,\vx,\vt)$ respectively given $a$ and $b$. If a standard cumulative product is used to precalculate values of $\prod_{k=j}^{j+c}p_{\phi}(z_k|t_i)$, for all values of $j$ and $c$, then calculating all necessary posterior marginals can be done in $\mathcal{O}(L\cdot M^2)$ time. In our implementation, we use automatic differentiation on the dynamic program for $a(L,M)$ to calculate all necessary gradients.

\section{Empirical Protocols}

\paragraph{Datasets:}

To evaluate the predictive performance of our model, we used two real mobile health datasets: mPuff \cite{ali2012mpuff} and puffMarker \cite{saleheen-ubicomp2015}. Both are smoking detection datasets and were gathered using similar protocols. Each subject wore a wireless respiration chest band sensor. Subjects were asked to smoke a cigarette while an observer recorded each smoking puff by pressing a button in a smartphone application. The resulting respiration time series were discretized by segmenting them into respiration cycles (a single inhalation and exhalation) and features were calculated on each respiration cycle \cite{saleheen-ubicomp2015}. Base features include inhalation and exhalation durations and amplitudes. We calculated further histogram based transformations of these features to allow for non-linearities. We note that while the discretization and featurization steps affect end-to-end classification performance, they are orthogonal to the main contribution of the paper. 
%

\begin{tabular}{|c|c|c|c|c|}\hline
	Dataset & Subjects & Sessions & \# Pos & \# Neg\\\hline\hline
	mPuff & 4 & 13 & 179 & 3180 \\\hline
	puffMarker & 6 & 30 & 305 & 7623 \\\hline
\end{tabular}

The mPuff and puffMarker datasets are organized into sessions that each contain a single smoking episode surrounded by different amounts of non-smoking activity. Basic information about the datasets is presented in the table above. Both datasets exhibit heavy class imbalance, which is typical in many mHealth settings. In our experiments, we consider the start of each respiration cycle to be the instance timestamp. Both data sets had the originally recorded positive event timestamps manually realigned with the instance sequence 
using a visualization procedure. We have access to these manually aligned labels for both data sets, which we consider to be the ground truth instance labels.

To test the behavior of our model under different amounts of observation noise, we generated synthetic observation timestamps, $\vz$, using the following noise model, which conditions on the ground truth instance timestamps and labels:
\begin{align}
	p(o_i=1|y_i=1) &= 1 - p(o_i=1|y_i=0) = \pi\\
	p(o_i>1|y_i=1) &= p(o_i>0|y_i=0) = 0\\
	p(z_j|t_i) &= \mathcal{N}(z_j;t_i,\sigma^2)
\end{align}
We varied $\pi \in (0,1)$ and $\sigma \in \mathbb{R}^+$ to simulate different amounts of missed/spurious observations and timestamp noise respectively. 

The puffMarker dataset also includes the original noisy event time stamps as recorded during the observation sessions. This allows us to conduct experiments where we train on real observation timestamps, but test on held out instances with ground truth labels provided by the manual alignment process. We chose a normal noise distribution for the synthetic observation process because it matches the noise distribution observed in the real puffMarker data, as shown in Figure 1 in the supplementary material.
%

\paragraph{Models:}

In the following experiments, we evaluate the proposed framework with a range of modeling choices. We evaluated three different base-classifier options: logistic regression (LR-M), kernel logistic regression with a radial basis function (RBF) kernel (KLR-M), and a small feedforward neural network (NN-M). For each classifier, we placed a zero-mean Gaussian prior with tuned variance on the classifier parameters $\theta$. As our observation count model, we used a simple Bernoulli distribution $p_\pi(o_i=1|y_i) = 1-p_\pi(o_i=0|y_i) = \pi_{y_i}$ for $\pi_v \in (0,1)$. This allows the model to account for both false positives and false negatives. In this case, the maximum number of observations generated by an instance is one, so the complexity of inference reduces to $\mathcal{O}(L\cdot M)$. We placed a beta prior with tuned parameters on $\pi$. Finally, we used the following gaussian mixture distribution as our event timestamp noise model:
\begin{align}
	p_\phi(z_j|t_i) = \sum_{k=1}^K \gamma_k \mathcal{N}(z_j; t_i + \mu_k, \sigma_k^2)
\end{align}
where $\phi = \{\mathbf{\mu},\mathbf{\sigma},\mathbf{\gamma}\}$. We placed a standard normal prior on each $\mu_k$ and an inverse-Gamma prior with shape $\alpha=1$ and scale $\beta=1$ on each $\sigma_k^2$. In our synthetic experiments we fixed $K=1$ so the observation noise model reverted to a single Gaussian; however, in our experiments on real data, we evaluated a range of $K$ values to allow for more complex noise distributions.


We compared these instances of our proposed framework with logistic regression models trained using different transformations of the available instance and positive event label sequences. In particular, we evaluated the logistic regression model learned with positive event labels manually aligned to instances (LR-H). If the manual alignment process contains no errors, these experiments can be thought of as producing best-case results given the discretization and featurization selected. We also evaluated the naive supervision strategy where we assign a positive event label to the instance that is closest in time to each noisy positive event time stamp (LR-N). This approach simply treats the noisy event time stamps as if they were correct. 

We also compared our model to a traditional multi-instance learning strategy that produces an instance-level classifier (MI). In particular, we adapted the MI-SVM training algorithm from \cite{andrews2002support} to train logistic regression models. This algorithm works by forming \emph{bags} of instances which are labeled as either positive or negative. A negative bag label indicates that the bag contains only negative instances and a positive bag label indicates that at least one of the instances is positive. For a given bag size, $B$, we formed bags by segmenting the base sequence into non-overlapping segments of length $B$. We generated bag labels by applying the naive alignment strategy described above and labeling a bag as positive if at least one positive event time stamp fell inside of the bag. 

The MI-SVM algorithm alternates between picking a representative instance from each positive bag, called a \emph{witness}, and training a classifier on the witnesses plus the negative instances. The witnesses are chosen to minimize the classifier's loss function. To our knowledge, this strategy has never been applied to logistic regression. In the case of logistic regression, this alternating algorithm results in a non-decreasing objective and therefore converges to a local optima. We tuned both the bag size $B$ and the classifier regularization hyper-parameters.

\paragraph{Train and Test Procedures:}

We evaluated all models using a 10-fold cross-validation procedure where the folds were generated across sessions. We tuned all hyper-parameters to maximize $F_1$ score on hand-aligned labels using a further 10-fold cross-validation procedure on the training set. The regularization parameters for each model were tuned across an exponential grid while the bag size for the multi-instance models was tuned on a linear grid. Results for all models are presented in terms of $F_1$ score which was chosen due to the heavy class imbalance and to highlight the performance on the class of interest, smoking puffs. Performance results were averaged across all folds. For synthetic data, results were further averaged across synthetic observations generated with two different seeds.

\section{Experiments and Results}

\paragraph{Experiment 1: Performance Under Varying Noise Conditions}

\begin{figure*}[t!]
        \begin{subfigure}{0.33\textwidth}
                \centering
                \includegraphics[width=\linewidth]{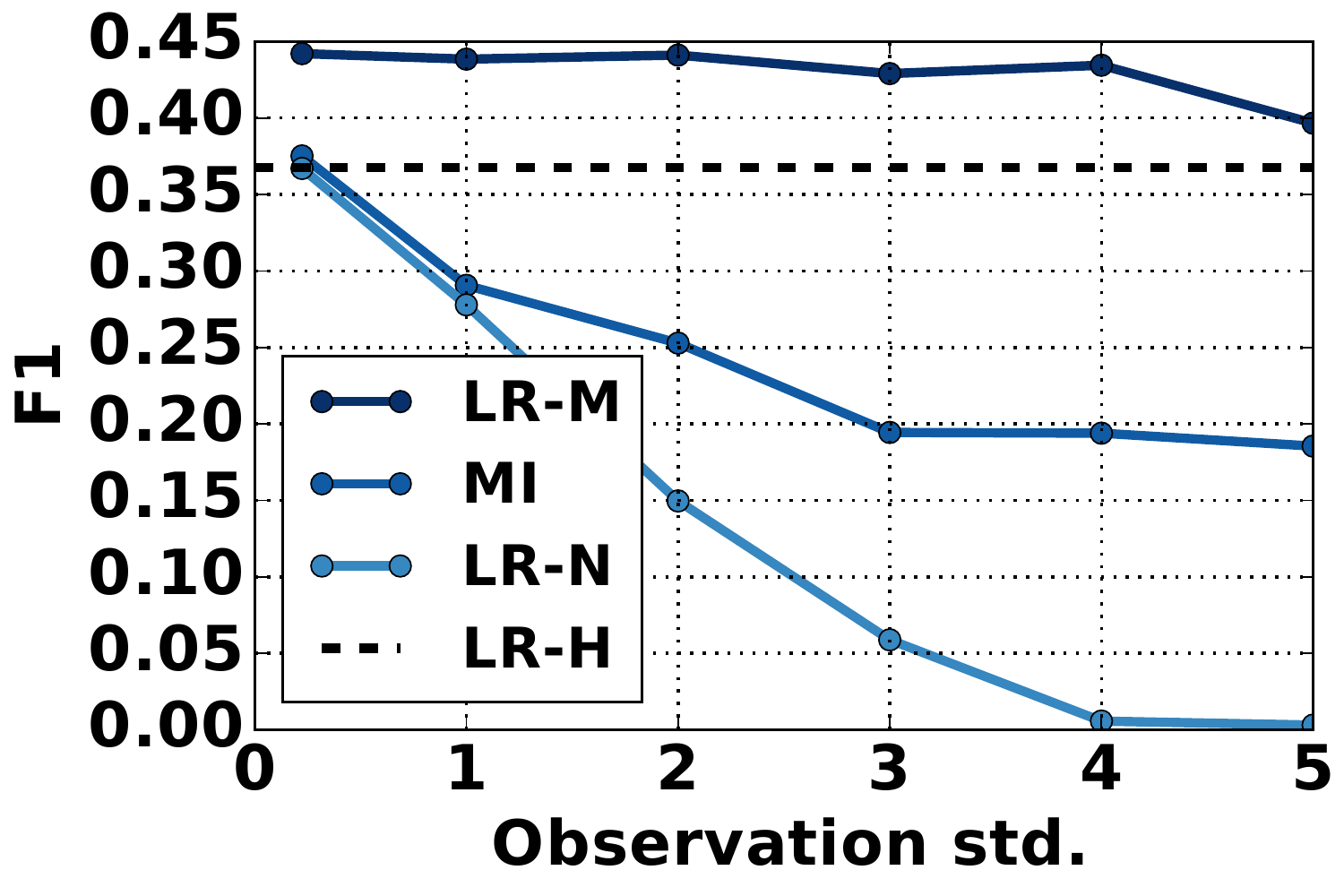}
                \caption{}
        \end{subfigure}
        \begin{subfigure}{.33\textwidth}
                \centering               
                
				\includegraphics[width=\linewidth]{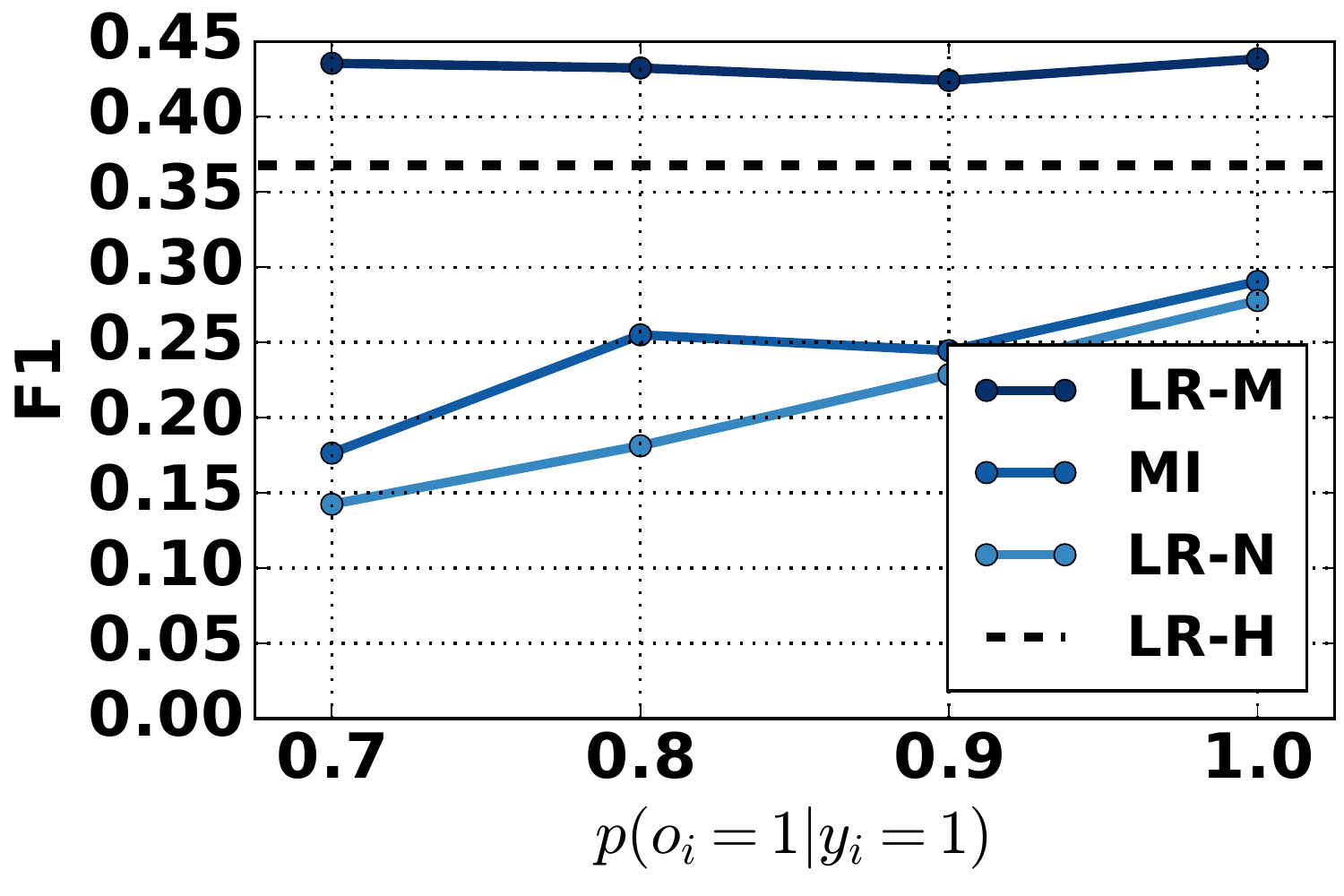}
                \caption{}
        \end{subfigure}
        \begin{subfigure}{.33\textwidth}
                \centering
				\includegraphics[width=\linewidth]{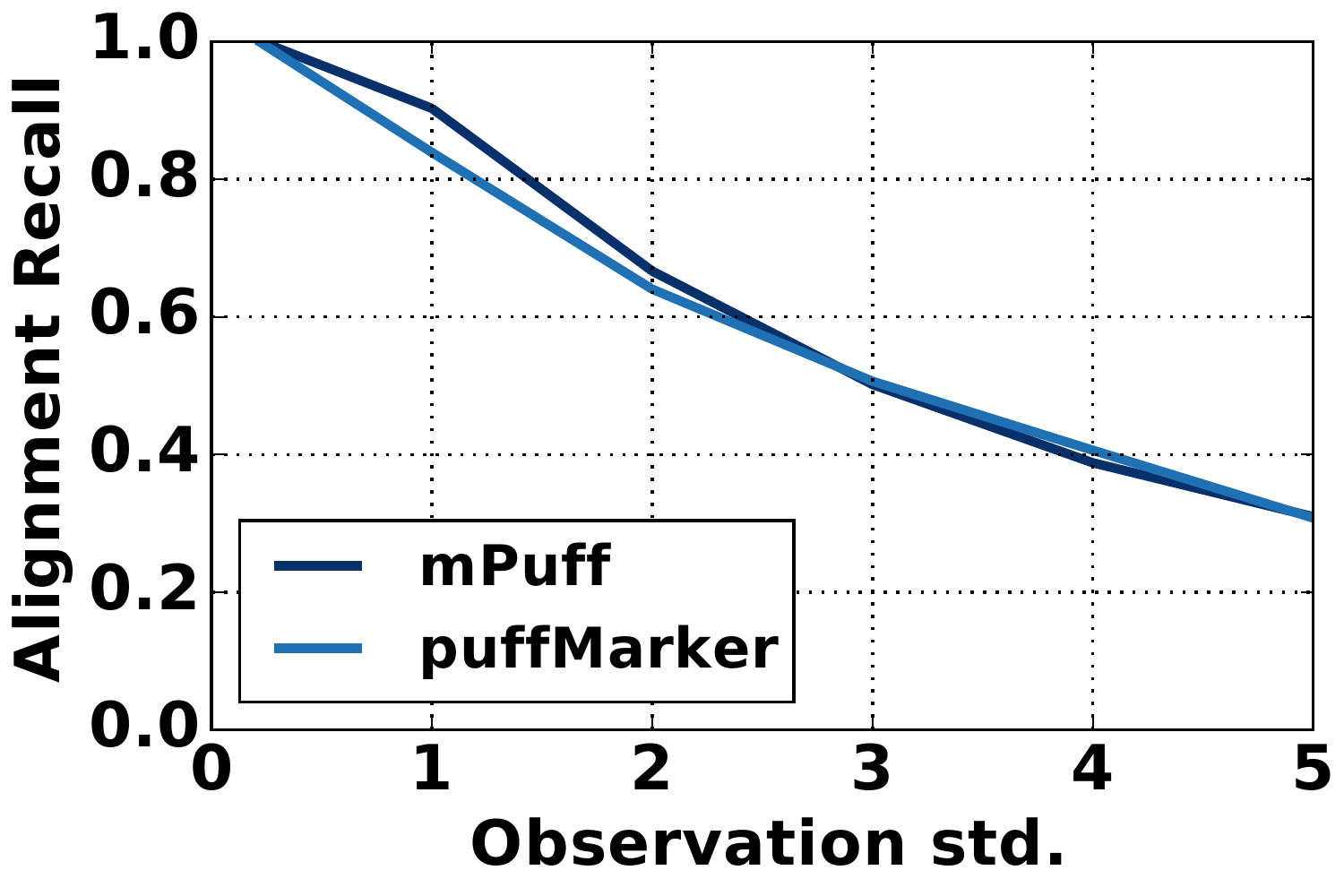}
                \caption{}
        \end{subfigure}\\
        \begin{subfigure}{0.33\textwidth}
                \centering
                \includegraphics[width=\linewidth]{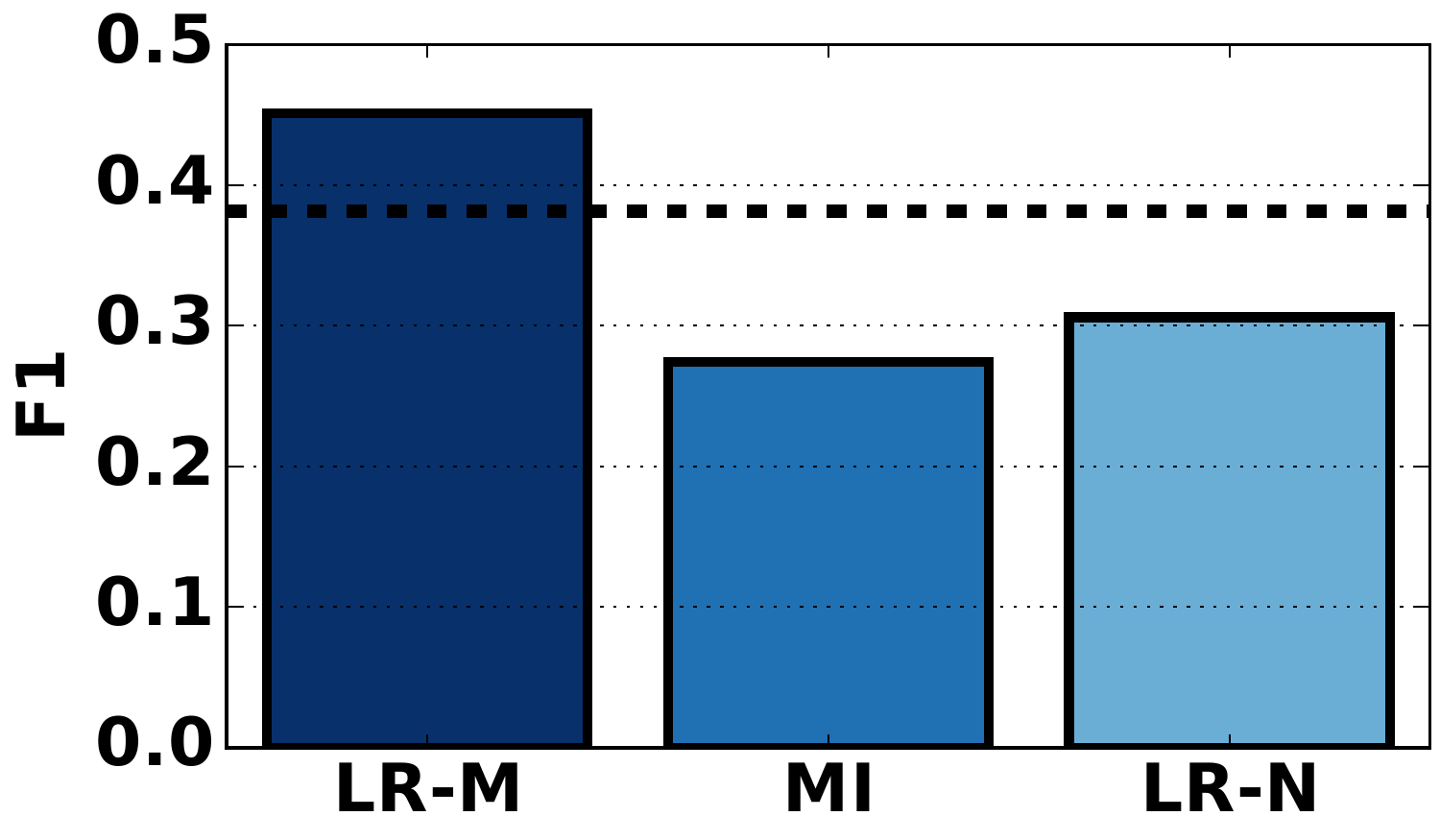}
                \caption{}
        \end{subfigure}
        \begin{subfigure}{.33\textwidth}
                \centering               
                \includegraphics[width=\linewidth]{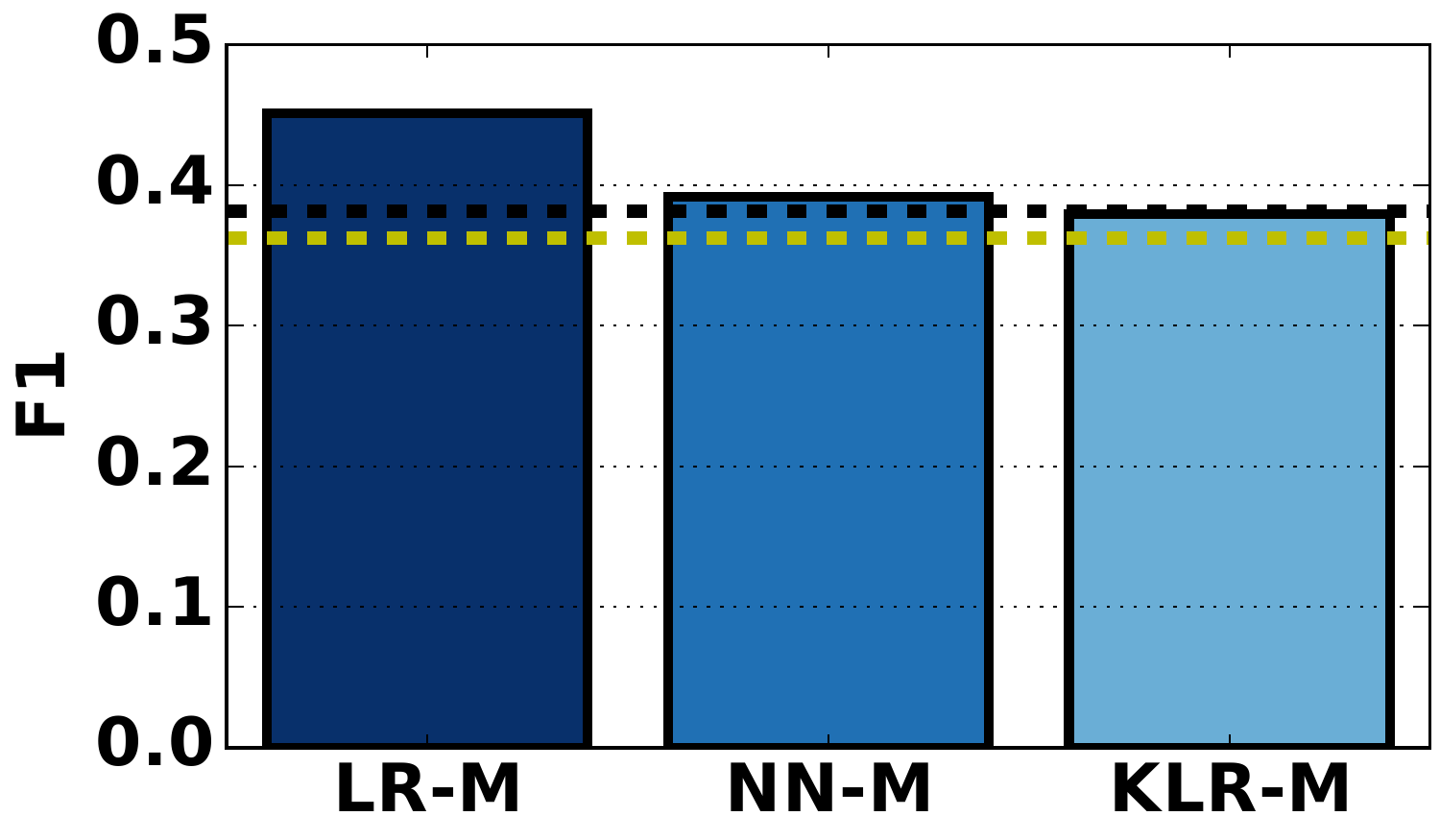}
                \caption{}
        \end{subfigure}
        \begin{subfigure}{.33\textwidth}
                \centering
			    \includegraphics[width=\linewidth]{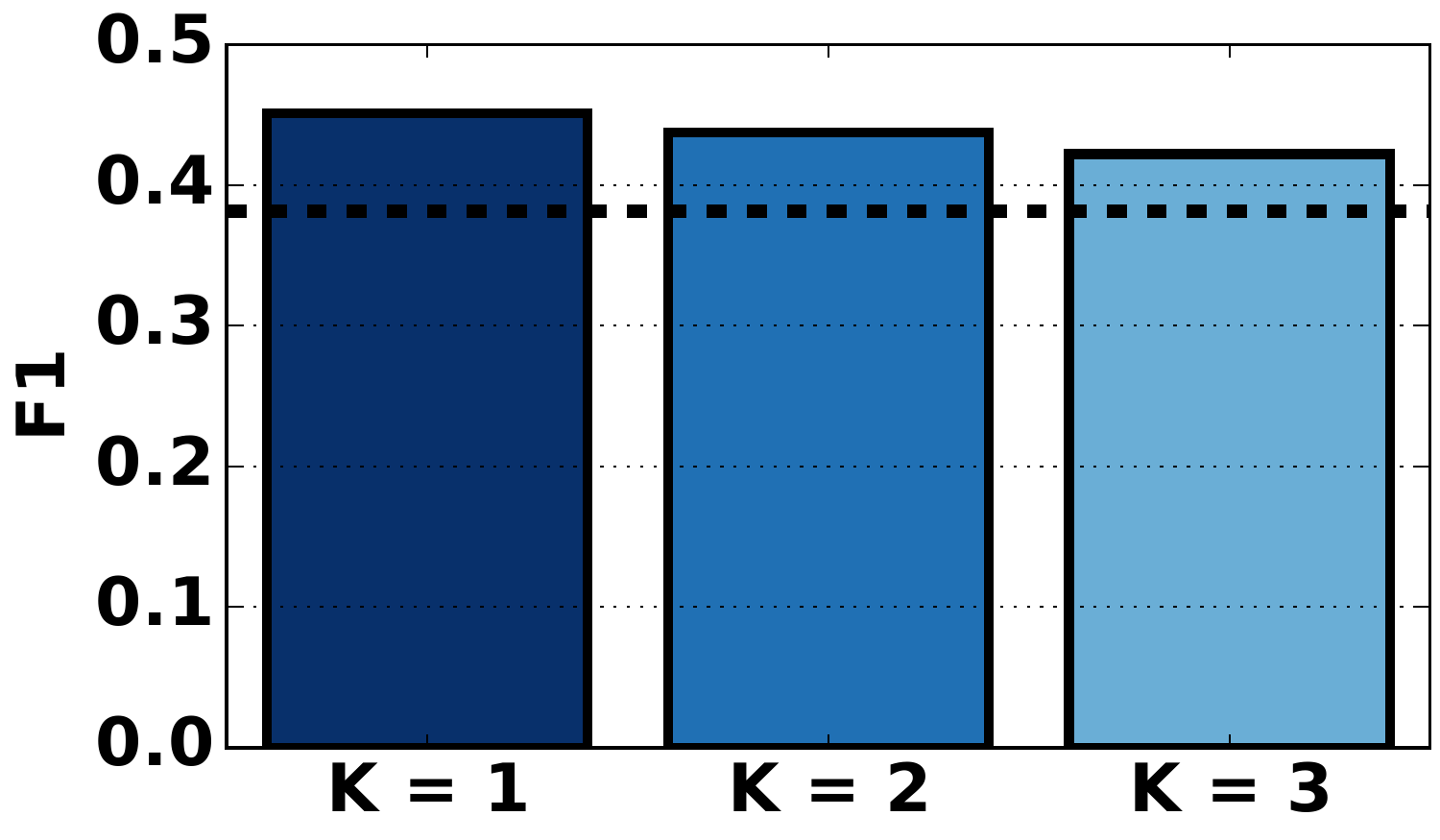}
                \caption{}
        \end{subfigure}
		\caption{Figures (a) and (b) show the prediction performance for all logistic regression-based models when varied amounts of synthetic noise are added to the hand aligned labels of the mPuff dataset. In Figure (a), we vary the standard deviation of the timestamp noise and in Figure (b), we vary the probability of observing a positive instance. Figure (c) shows the recall of the labels generated by the naive alignment strategy. Figures (c), (d), and (e) show the performance of various models on the real puffMarker data. Figure (e) shows the perfomance of the LR-M model using a GMM noise model with different numbers of components. In all plots, the black dashed line corresponds to LR-H and the yellow dashed line corresponds to NN-H.}
        \label{fig:synthetic}
\end{figure*}

We evaluated how robust the proposed model is to observation noise by testing the predictive performance of the logistic regression-based models on the mPuff and puffMarker datasets with synthetic observation timestamps. Figure \ref{fig:synthetic} (a) shows the performance in terms of $F_1$ of these models on the puffMarker dataset with varied observation noise, but fixed observation count noise. For this experiment, we varied the standard deviation ($\sigma$) of the synthetic noise from $0.25$ to $5.0$ with the observation count probability ($p(o_i=1|y_i = 1) = \pi$) fixed at $1$. While the performance of all models degrades as the amount of noise increases, the performance of the proposed model (LR-M) degrades noticeably slower than the traditional multi-instance method (MI), which in turn only slightly outperforms the naive alignment strategy (LR-N) on this task. To understand why the performance of the naive and multi-instance methods degrade so much, Figure \ref{fig:synthetic} (c) shows the proportion of true positive labels that are assigned positive labels by the naive alignment procedure on the mPuff and puffMarker datasets across the same range of noise standard deviations. Even at $\sigma = 2.0$, only $65\%$ of true positives retain a positive label under the naive supervision strategy. As expected, the performance of the baseline methods track this plot very closely. 


Figure \ref{fig:synthetic} (b) shows the performance of the logistic regression based models when $\sigma$ is kept fixed at 1, but $\pi$ is varied between $0.7$ and $1$. $\sigma = 1$ was chosen because it matches the variance of the empirical noise distribution of the puffMarker dataset. Again, the performance of the multi-instance and naive alignment methods degrades much more quickly than the proposed method as noise is added. We obtained similar results for both experiments on the mPuff dataset and the mPuff results can be found in the supplemental materials.


%

\paragraph{Experiment 2: Performance on Real Timestamps}

We evaluated the performance of the logistic regression-based models on the puffMarker dataset using the original positive event timestamps. The predictive performance in terms of $F_1$ is shown in Figure \ref{fig:synthetic} (d). The LR-M model substantially outperforms both the multi-instance and naive methods. The improvement of LR-M over LR-N and MI is statistically significant at the $p=0.005$ level using a paired t-test with Bonferroni correction. As in the synthetic experiments on the puffMarker dataset, the LR-M model outperforms a logistic regression model trained on the hand aligned labels. One possible explanation for this behavior is that there is a non-trivial amount of annotation error that our model is correcting for. Another possible explanation for this result is that the there is non-trivial amounts of class overlap in the feature space. Our model allows for the possibility that there are false negatives in the labels and so it is able to get around this overlap by treating positive looking negative instances as positive, whereas the other models must treat them as negative. This is supported by the observation that the precision of our model is slightly lower than the two baselines, but the recall is much higher.

Finally, in order to demonstrate the flexibility of the proposed framework, we tested alternative noise models and base classifiers. Figure \ref{fig:synthetic} (e) shows the performance of the proposed framework on real puffMarker data using three different base classifiers: logistic regression (LR-M), kernel LR with an RBF kernel (KLR-M), and a feed-forward neural network (NN-M). LR-M outperforms both of the non-linear models, though all models perform as well as a classifier trained directly on hand-aligned labels. Figure \ref{fig:synthetic} (f) shows the predictive performance of the LR-M model as the number of components $K$ in the GMM timestamp noise distribution is increased. Performance degrades slightly as the flexibility of the noise distribution is increased. One possible explanation for this behavior is that a normal distribution is already a good fit to the data, and the increased number of components leads to overfitting.

\section{Conclusions and Future Work}

In this work, we have addressed the problem of labeling discretized time series when only noisy positive event timestamps are available. The primary contributions of this paper are a novel framework for learning descrimitive classifiers in this setting based on maximizing the marginal likelihood of a latent variable model. This framework admits efficient exact inference and can support a broad class of probabilistic discriminative classifiers. We presented results showing that the proposed framework substantially outperforms multiple baselines under moderate to severe observation noise conditions and on a real mHealth dataset.

One possible future direction is to extend this framework to more complex or structured base classifiers and more sophisticated observation models. For example, the same learning framework can be used unchanged with a recurrent neural network and requires only small modifications to incorporate a linear chain conditional random field as the base classifier. The framework can be further extended to models that reason about higher level structure, such as segmentations of the time series, integrating over uncertainty at these levels as well. In mobile health in particular, we often have noisy self-reported observations that tell us only a rough period in which positive labels may have occurred.

\section*{Acknowledgments} 
The authors would like to thank members of the MD2K Center (\url{http://www.md2k.org}) for helping to enable this research. This work was partially supported by the National Institutes of Health under award 1U54EB020404, and the National Science Foundation under award IIS-1350522.

\bibliographystyle{plain}
\bibliography{related_work_2}

\end{document}